# Diamond Abrasive Electroplated Surface Anomaly Detection using Convolutional Neural Networks for Industrial Quality Inspection

Parviz Ali

https://www.linkedin.com/in/parviz-ali-b7549010/

parviza9999@gmail.com

**Abstract**

Electroplated diamond abrasive tools require nickel coating on a metal surface for abrasive bonding and part functionality. The electroplated nickel-coated abrasive tool is expected to have a high-quality part performance by having a nickel coating thickness of between 50% to 60% of the abrasive median diameter, uniformity of the nickel layer, abrasive distribution over the electroplated surface, and bright gloss. Electroplating parameters are set accordingly for this purpose. Industrial quality inspection for defects of these abrasive electroplated parts with optical inspection instruments is extremely challenging due to the diamond's light refraction, dispersion nature, and reflective bright nickel surface. The difficulty posed by this challenge requires parts to be quality inspected manually with an eye loupe that is subjective and costly. In this study, we use a Convolutional Neural Network (CNN) model in the production line to detect abrasive electroplated part anomalies allowing us to fix or eliminate those parts or elements that are in bad condition from the production chain and ultimately reduce manual quality inspection cost. We used 744 samples to train our model. Our model successfully identified over 99% of the parts with an anomaly.

Keywords: Artificial Intelligence, Anomaly Detection, Industrial Quality Inspection, Electroplating, Diamond Abrasive Tool

## Introduction

An anomaly is an event or item that deviates from what is expected. An anomaly's frequency is low compared to standard events' frequency. The anomalies that can occur in the electroplated parts are usually random, some examples are changes in color or surface texture, scratches in the nickel layer, abnormal abrasive distribution, missing abrasive, or incorrect abrasive exposure related to nickel layer thickness.

Anomaly Detection allows us to fix or eliminate those parts or elements that are in bad condition from the production chain. Safety-relevant products in the medical or automobile industry require to realize a 100% quality inspection. Aside from suitable measurement techniques, this requires appropriate algorithms, as manual inspection is not only subjective, repetitive, and prone to human error but often also infeasible with production rates of multiple parts per second (Flosky and Vollertsen 2014). Algorithms for automatic surface inspection are to a large part based on manually engineered features (Neogi, Mohanta, and Dutta 2014), most commonly statistical and filter-based (Xie 2008). While the introduction of expert knowledge often allows for the creation of powerful features, this process is laborious and might be necessary for each new product with multiple abrasive sizes. General solutions that can automatically adapt to new problem sets could hence yield significant time and cost advantages. Anomaly detection, in factories, is a useful tool for Quality Control Systems because of its features and is a big challenge for Machine Learning Engineers. With the use of vision-based generalized artificial intelligence anomaly detection, manufacturing costs are reduced because of the avoidance of producing and marketing defective products.

Using Supervised Learning is not a recommended practice because of: the need for intrinsic features in anomaly detection and the use of the low quantity of anomalies in a full dataset (training/validation). On the other hand, image comparison could be a feasible solution but Standard Images handle several variables such as light, object position, distance to object, and others; which does not allow the pixel-to-pixel comparison with a standard image. Pixel-to-pixel comparison is integral in the detection of anomalies.

One such solution is convolutional neural networks (CNN) (see figure 1). They have become the driving factor behind many recent innovations in the field of computer vision and allowed significant advances in various applications, such as object classification (Krizhevsky, Sutskever, and Hinton 2017) or semantic image segmentation (Long,

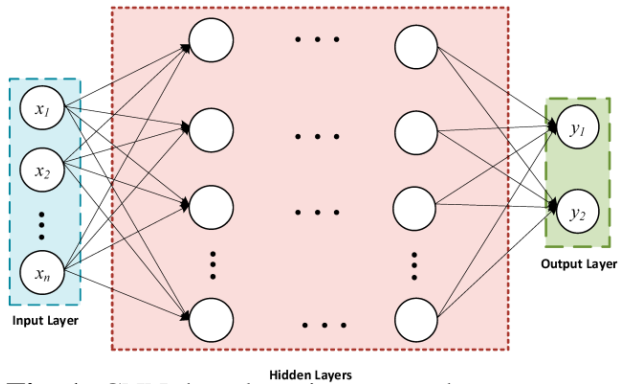

**Fig. 1:** CNN deep learning network

Shelhamer, and Darrell 2015). CNNs have also recently been successfully applied to industrial surface inspection (Weimer, Scholz-Reiter, and Shpitalni 2016). A prerequisite for the training of CNNs is the availability of a sufficiently large body of training data. Depending on the intra-class variance of the different defect types and non-defective areas, this could mean hundreds or up to several thousand samples. However, with well-optimized processes, there is often an abundance of non-defective samples while the availability of defect samples is very limited. One solution to this problem is to shift the training objective from defect classification toward anomaly detection, as such an approach would require no defective samples for training. Another potential benefit is that a well-performing anomaly detection algorithm would also be able to detect previously unknown defect classes, i.e. constitute a more general solution to the quality inspection problem.

## Literature Review

A variety of approaches currently exist for anomaly detection that can roughly be categorized into probabilistic, reconstruction-based, domain-based, information-theoretic, and distance-based (Pimentel et al. 2014). Recently there have been approaches for using features learned by CNNs for surface defect detection (Natarajan et al. 2017) and (Napolitano, Piccoli, and Schettini 2018). Both techniques rely on transfer learning, i.e. learning features by solving a different task and using these features for surface defect detection. However, the first of these approaches, presented by Natarajan et al. (2017) still requires defective samples for training and therefore does not solve the anomaly detection problem as stated in this work. The approach closest to our work is a method introduced by Napoletano et al. (2018) for anomaly detection in nanofibrous materials. They determine the similarity between image patches based on features of a CNN that they trained for object classification on the ILSVRC 2015 ImageNet data set. Anomalies are thereby detected by their large distance to normal regions in the feature space. However, in contrast to our approach the CNN features are not trained specifically to assign low distance to similar parts and high distance to dissimilar parts, i.e. learn a similarity metric. Meaningful distances in the feature space are therefore a mere byproduct of solving the classification task and not explicitly sought. Changing the objective towards explicitly learning a similarity metric is hence a promising direction for further improvement. Here we present a solution for learning meaningful features for distance-based surface anomaly detection using a deep learning network.

Deep metric learning uses deep neural networks to directly learn a similarity metric, rather than creating it as a byproduct of solving e.g. a classification task. They are especially useful for solving tasks where the amount of object classes is possibly endless and the classification framework is not feasible anymore. Popular applications are hence image retrieval (Song et al. 2016), person re-identification (Herman, Beyer, and Leibe 2017) or face verification (Schroff, Kalenichenko, and Philbin 2015). One popular architecture for deep metric learning is triplet networks (Hoffer and Ailon 2015). In this setup triplets consisting of three different images are fed into the same network. Two of these images belong to the same class and one image belongs to a different class. The network is then trained to create a feature space in which same-class samples have a lower distance to each other than to samples from other classes.

**Data**

Data Collection

We collected 930 sample images of 45-micron size diamond electroplated on brushed stainless-steel cards (see figure 2) with 0.0625-inch thickness. The parts were 2 inches high and 3.25 inches wide. A total of 474 good part samples and 456 bad part sample images were captured using Plugable USB Digital

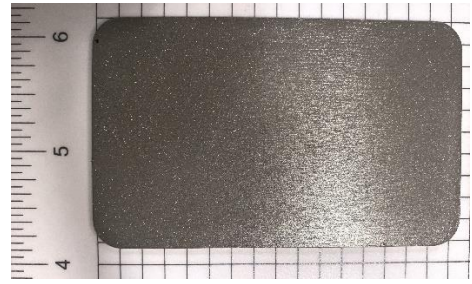

**Fig. 2:** Diamond electroplated card

Microscope with 250x magnification and 4 integrated led lamps, set at 50% intensity, connected to a Portable Computer with Windows 11 operating system (see figure 3). Plugable Digital Viewer software was used to capture 1600 x 1200-pixel color images one part at a time using a manual trigger. The fixed stand microscope was placed on a flat granite stone for uniform lighting.

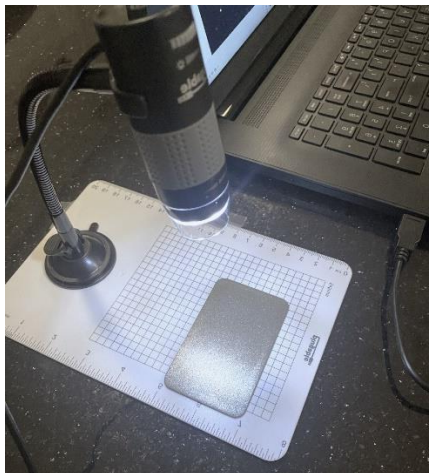

**Fig. 3:** Data collection setup

## Data Preprocessing

Captured color images were converted to grayscale and reduced to the size of 300 x 300 pixels. The auto serialized tagged images were saved in subfolders labeled 'bad' and 'ok'. Training and test data were saved in separate subfolders (see figure 4).

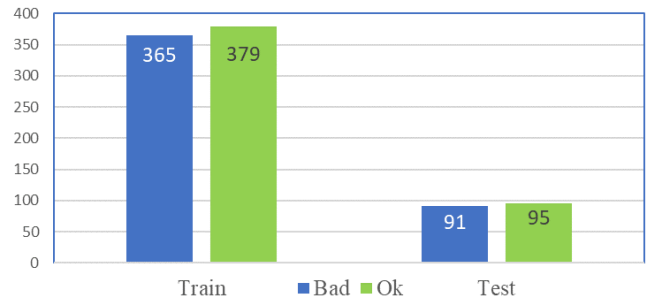

**Fig. 4:** Data Distribution

## Method

We trained our CNN model to classify the 300 x 300-pixel size grayscale images of electroplated parts. The CNN model is used as an automatic feature extractor from the images so that it can learn how to distinguish between bad and ok electroplated parts. The model effectively uses the adjacent pixel to downsample the image and then uses a prediction (fully connected) layer to solve the classification problem (see figure 5). We used a training set of 632 examples, a validation set of 112 examples, and a test set of 281 examples, comprising 186 'bad' and 95 'ok' sample images (see figure 6). We apply the on-the-fly data augmentation technique to expand the training dataset size by creating a modified version of the original image to improve model performance and the ability to generalize. Image augmentation parameters include; the degree range for random rotations set at 90 degrees, fraction range of the total width to be shifted, fraction range of the total height to be shifted, degree range for random shear in a counter-clockwise direction, fraction range for random zoom, random vertical and horizontal image flip, fraction range for picking a brightness shift value, rescale the pixel values to be in ranges 0 and 1, reserve 15% of the training data for validation, and the rest 85% for model fitting.

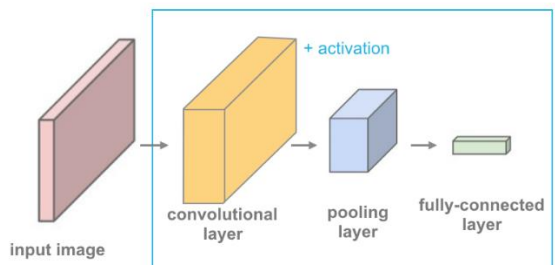

**Fig. 5:** CNN model

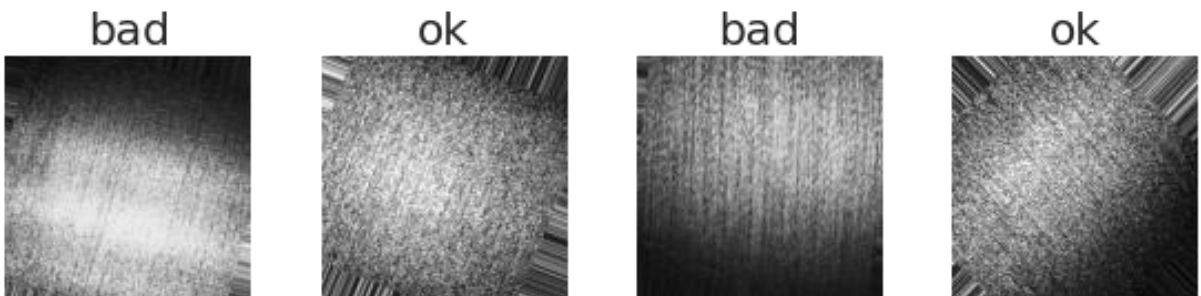

**Fig. 6:** Training images with data augmentation

    The first convolutional layer of our model consists of 32 filters with kernel size matrix 3 by 3. Using 2-pixel strides at a time we reduce the image size by half. The first max-pooling consists of matrix 2 by 2 (pool size) and 2-pixel strides at a time to further reduce the image size by half. The second convolutional layer is similar to the first convolutional layer with 16 filters only. The flattening layer is used to convert two-dimensional pixel values into one dimension and fed into the fully-connected layer. The first dense layer and dropout have 128 units and 1 bias unit. A dropout rate of 20% is used to prevent overfitting. The second dense layer and dropout have 64 units and 1 bias unit. A dropout rate of 20% is also used to prevent overfitting. The output layer consists of only one unit and activation is a sigmoid function to convert the scores into a probability of an image is of the bad part. We use rectified linear unit (ReLU) activation function for every layer except the output layer.

**Results**

The 25 pixels by 25 pixels image (see figure 7) shows a random pattern of the part. This image does not reveal a visual pattern indicating the uniqueness of each part. Evaluation of the training results (see figure 8) reveals the divergence between loss decrease and the increase in accuracy.

The model achieved 61% accuracy and 79% validation accuracy after running for 10 epochs. The model then started overfitting. After running for 20 epochs, the model achieved 70% accuracy and 87% validation accuracy.

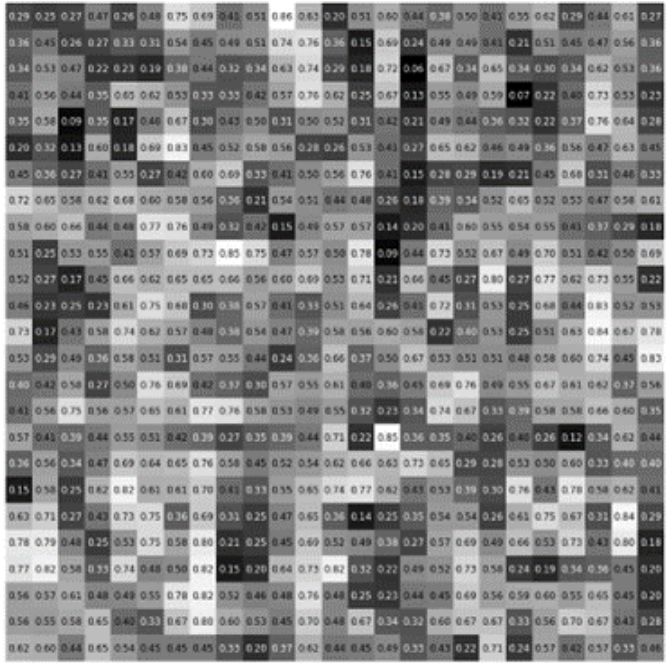

**Fig. 7:** 25 x 25 pixels image of the plated part

Out of 95 good sample parts, the model correctly predicted 94 parts being ok. Only one part was misclassified as being a bad part. Of the bad part sample, the model accurately predicted all of them to be a bad part (see table 1).

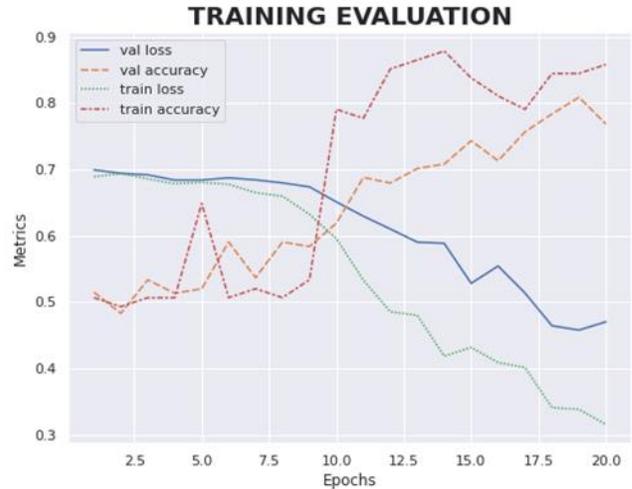

**Fig. 8:** Training evaluation

Table 1. Model test data precision, recall score

|  | precision | recall | f1-score | support |
|---|---|---|---|---|
| 0 | 1.0000 | 0.9895 | 0.9947 | 95 |
| 1 | 0.9891 | 1.0000 | 0.9945 | 91 |
| accuracy |  |  | 0.9946 | 186 |
| macro avg | 0.9946 | 0.9947 | 0.9946 | 186 |
| weighted avg | 0.9947 | 0.9946 | 0.9946 | 186 |

|  |  | Predicted | |
|---|---|---|---|
|  |  | ok | bad |
| Actual | ok | 94 | 1 |
|  | bad | 0 | 91 |

**Discussion**

As expected, due to the diamond's light refraction and dispersion nature, the pixelated image of the part is random without a visible pattern. The CNN model starts to overfit after running for 10 epochs as the loss decreases drastically.

Even with image augmentation, generalization is difficult as we notice the widening gap between the training and validation loss. All of our defective sample parts are from the same production run causing the defect to be similar. Also, the number of sample parts available for training is low ultimately affecting model generalization and causing the model to overfit rapidly.

Our goal was to minimize the case of False Negative, where the bad part is misclassified as ok. This can cause the whole order to be rejected and create a big loss for the company. Therefore, in this case, we prioritized Recall over Precision. But if we take into account the cost of re-plating a product, we have to minimize the case of False Positive also, where the ok product is misclassified as bad. Therefore, we can prioritize the F1 score which combines both Recall and Precision. We notice a very high F1 score due to again, the low number of single production lot test samples available.

**Conclusions**

The model accurately identified 99% of the ok part (95) sample and 100% of the bad (91) part sample. We recommend using this model to study defective parts from multiple production run sources and to establish a more generalized model output. Further study should be done to determine if a correlation between process variables and defects exists. This will ultimately improve product quality and further reduce production quality inspection costs.

**Directions for future work**

This study has several limitations. We were not able to collect image data from multiple manufacturing production runs. This limited our ability to train the model to generalize product defects. Defective part samples are extremely difficult to gather in a short period of time in a highly controlled production run, we plan to extend our study over a longer span of time. We plan to study image data from multiple production batches to train the model for better generalization and accuracy. Manufacturing process setup variables need to be accounted for in future studies as production defects are highly dependent on the process setup and critical variables such as plate time, chemical concentration, anode placement, temperature, etc. based on our findings in this study. We would like to include these components when designing our future quality control AI model. Ultimately, our goal will be to use the results to fine-tune the process input variables based on any correlations discovered in the future study.

12

# Appendix

*Python code* (Morton 2020)*:*

## Import Libraries

As usual, before we begin any analysis and modeling, let's import several necessary libraries to work with the data.

```python
In [ ]: import os
# Data Analysis
import pandas as pd
import numpy as np
import keras

import tensorflow as tf
from tensorflow import keras
from tensorflow.keras import Model

from tensorflow.keras.utils import plot_model

# Visualization
import matplotlib.pyplot as plt
import seaborn as sns
sns.set()

# Neural Network Model
from keras.preprocessing.image import ImageDataGenerator
from keras.models import Sequential, load_model
from keras.layers import *
from keras.callbacks import ModelCheckpoint
from keras.utils.vis_utils import plot_model

# Evaluation
from sklearn.metrics import confusion_matrix, classification_report
```

## Load the Images

Here is the structure of our folder containing image data:

```
card_data
├──test
│   ├──bad
│   └──ok
└──train
    ├──bad
    └──ok
```

The folder `card_data` consists of two subfolders `test` and `train` in which each of them has another subfolder: `bad` and `ok` denoting the class of our target variable. The images inside `train` will be used for model fitting and validation, while `test` will be used purely for testing the model performance on unseen images.

## Data Augmentation

We apply on-the-fly data augmentation, a technique to expand the training dataset size by creating a modified version of the original image which can improve model performance and the ability to generalize. We will use with the following parameters:

- `rotation_range` : Degree range for random rotations. We choose 360 degrees since the product is a round object.
- `width_shift_range` : Fraction range of the total width to be shifted.
- `height_shift_range` : Fraction range of the total height to be shifted.
- `shear_range` : Degree range for random shear in a counter-clockwise direction.
- `zoom_range` : Fraction range for random zoom.
- `horizontal_flip` and `vertical_flip` are set to True for randomly flip image horizontally and vertically.
- `brightness_range` : Fraction range for picking a brightness shift value.

Other parameters:

- `rescale` : Eescale the pixel values to be in range 0 and 1.
- `validation_split` : Reserve 15% of the training data for validation, and the rest 85% for model fitting.

```python
from google.colab import drive
drive.mount('/content/drive')
```

Mounted at /content/drive

```python
train_generator = ImageDataGenerator(rotation_range = 90,
                                     width_shift_range = 0.05,
                                     height_shift_range = 0.05,
                                     shear_range = 0.05,
                                     zoom_range = 0.05,
                                     horizontal_flip = True,
                                     vertical_flip = True,
                                     brightness_range = [0.75, 1.25],
                                     rescale = 1./255,
                                     validation_split = 0.2)
```

We define another set of value for the `flow_from_directory` parameters:

- `IMAGE_DIR` : The directory where the image data is stored.
- `IMAGE_SIZE` : The dimension of the image (300 px by 300 px).
- `BATCH_SIZE` : Number of images that will be loaded and trained at one time.
- `SEED_NUMBER` : Ensure reproducibility.
- `color_mode = "grayscale"` : Treat our image with only one channel color.
- `class_mode` and `classes` define the target class of our problem. In this case, we denote the `bad` class as positive (1), and `ok` as a negative class.
- `shuffle` = True to make sure the model learns the bad and ok images alternately.

```python
IMAGE_DIR = '../content/drive/MyDrive/data/card_data/'
```

```python
IMAGE_SIZE = (300, 300)
BATCH_SIZE = 64
SEED_NUMBER = 123

gen_args = dict(target_size = IMAGE_SIZE,
                color_mode = "grayscale",
                batch_size = BATCH_SIZE,
                class_mode = "binary",
                classes = {"ok": 0, "bad": 1},
                seed = SEED_NUMBER)

train_dataset = train_generator.flow_from_directory(
                                        directory = IMAGE_DIR + "train",
                                        subset = "training", shuffle = True, **gen_args
validation_dataset = train_generator.flow_from_directory(
                                        directory = IMAGE_DIR + "train",
                                        subset = "validation", shuffle = True, **gen_a
```

```
Found 596 images belonging to 2 classes.
Found 148 images belonging to 2 classes.
```

We will not perform any data augmentation on the test data.

```python
test_generator = ImageDataGenerator(rescale = 1./255)
test_dataset = test_generator.flow_from_directory(directory = IMAGE_DIR + "test",
                                                  shuffle = False,
                                                  **gen_args)
```

```
Found 186 images belonging to 2 classes.
```

## Visualize the Image

We successfully load and apply on-the-fly data augmentation according to the specified parameters. Now, in this section, we visualize the image to make sure that it is loaded correctly.

### Visualize Image in Batch

Visualize the first batch (`BATCH_SIZE = 64`) of the training dataset (images with data augmentation) and also the test dataset (images without data augmentation).

```python
mapping_class = {0: "ok", 1: "bad"}
mapping_class
```

```
{0: 'ok', 1: 'bad'}
```

```python
def visualizeImageBatch(dataset, title):
    images, labels = next(iter(dataset))
    images = images.reshape(BATCH_SIZE, *IMAGE_SIZE)
    fig, axes = plt.subplots(8, 8, figsize=(16,16))

    for ax, img, label in zip(axes.flat, images, labels):
        ax.imshow(img, cmap = "gray")
        ax.axis("off")
        ax.set_title(mapping_class[label], size = 20)
```

```
        plt.tight_layout()
        fig.suptitle(title, size = 30, y = 1.05, fontweight = "bold")
        plt.show()

        return images
```

In [ ]: 
```
train_images = visualizeImageBatch(train_dataset,
                                   "FIRST BATCH OF THE TRAINING IMAGES\n(WITH DATA AUG
```

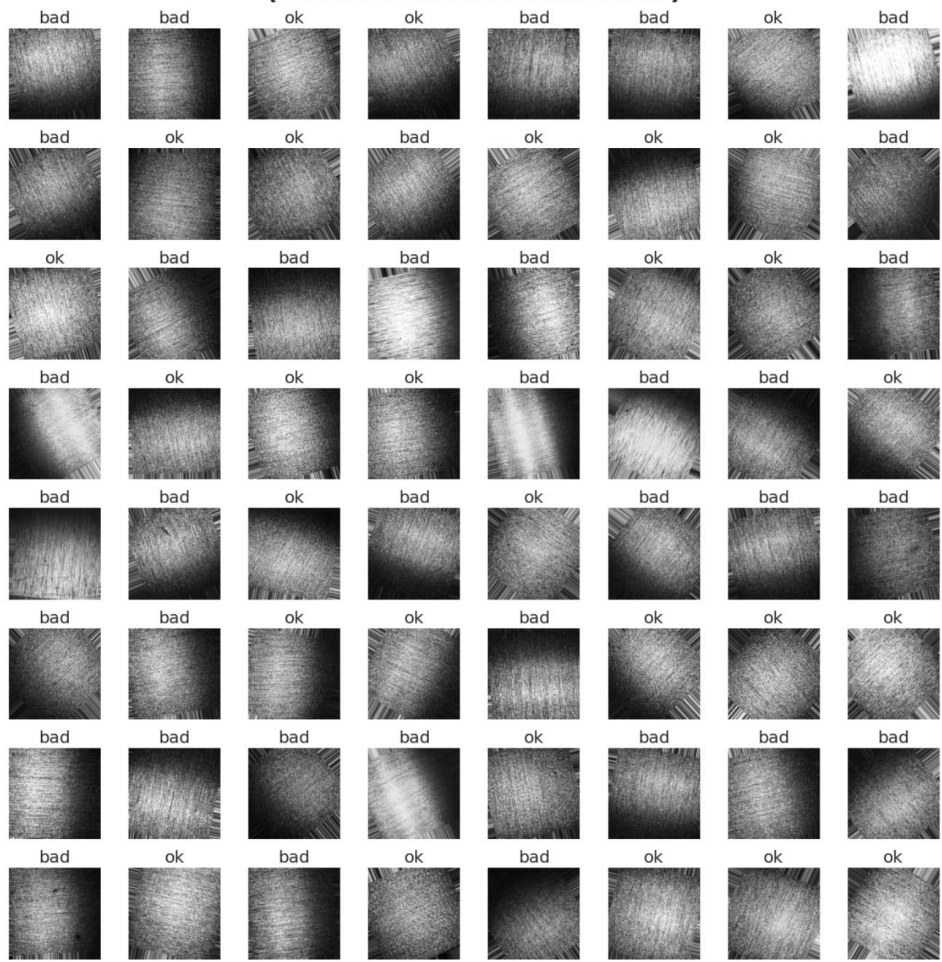

In [ ]:
```
test_images = visualizeImageBatch(test_dataset,
                                  "FIRST BATCH OF THE TEST IMAGES\n(WITHOUT DATA AUGME
```

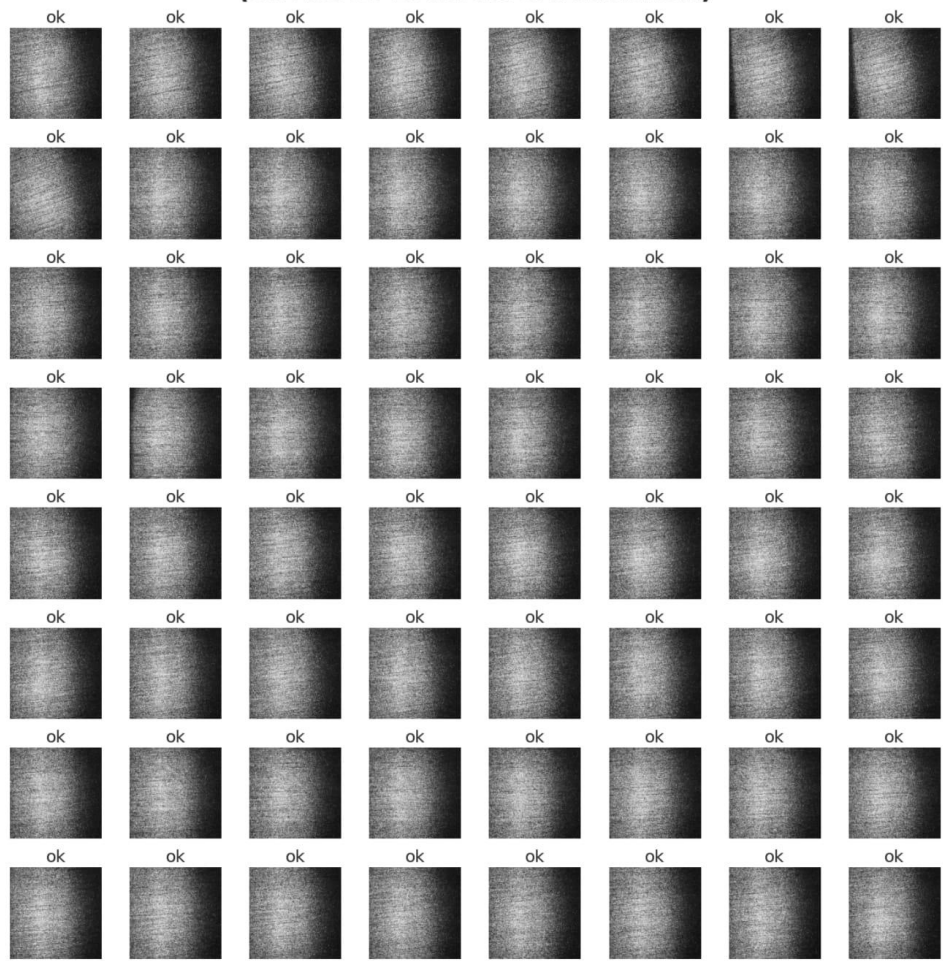

## Visualize Detailed Image

Let's also take a look on the detailed image by each pixel. Instead of plotting 300 pixels by 300 pixels (which computationally expensive), we take a small part of 25 pixels by 25 pixels only

```python
In [ ]:  img = np.squeeze(train_images[4])[75:100, 75:100]

         fig = plt.figure(figsize = (15, 15))
         ax = fig.add_subplot(111)
         ax.imshow(img, cmap = "gray")
         ax.axis("off")

         w, h = img.shape
         for x in range(w):
             for y in range(h):
                 value = img[x][y]
```

```
ax.annotate("{:.2f}".format(value), xy = (y,x),
            horizontalalignment = "center",
            verticalalignment = "center",
            color = "white" if value < 0.4 else "black")
```

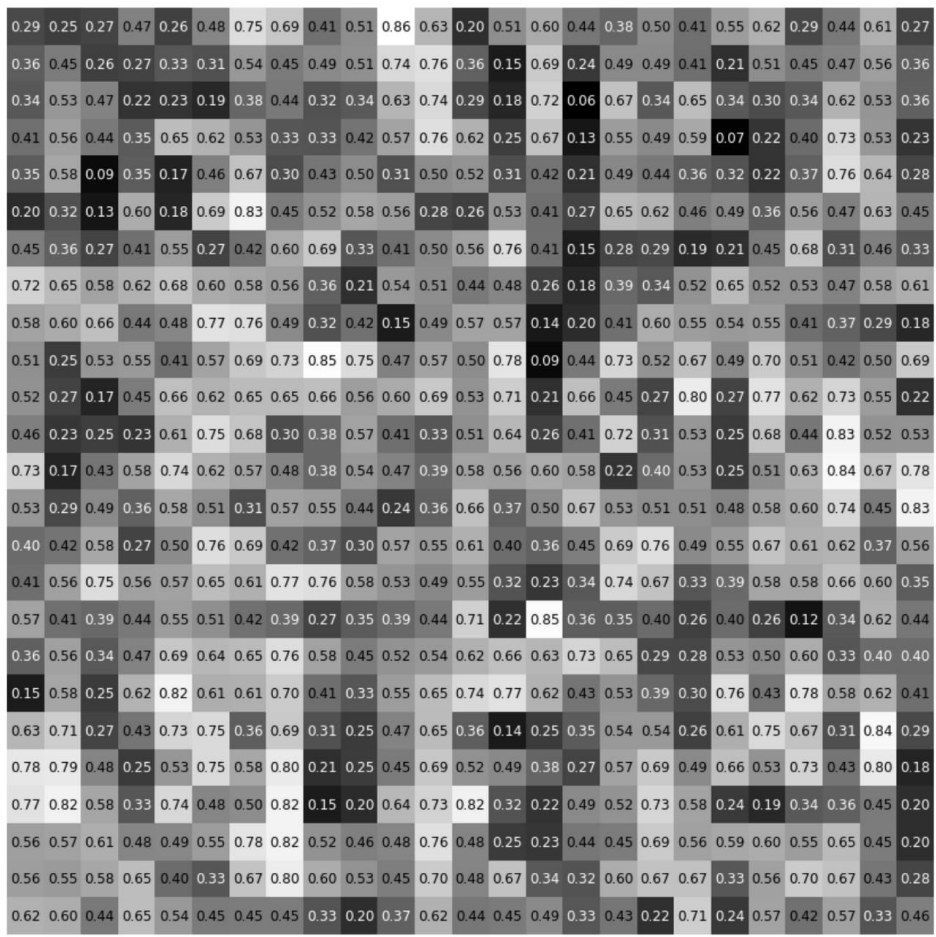

These are the example of values that we are going to feed into our CNN architecture.

## Training the Network

As mentioned earlier, we are going to train a CNN model to classify the casting product image. CNN is used as an automatic feature extractor from the images so that it can learn how to distinguish between `bad` and `ok` casted products. It effectively uses the adjacent pixel to downsample the image and then use a prediction (fully-connected) layer to solve the classification problem. This is a simple illustration by Udacity on how the layers are arranged sequentially:

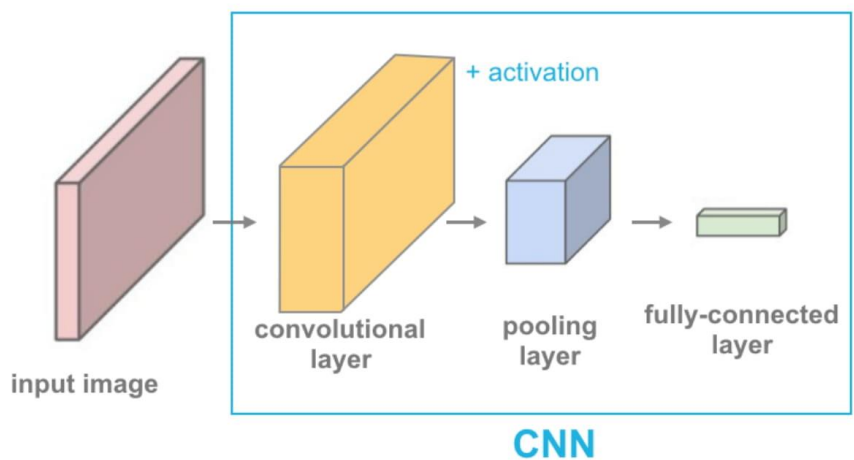

## Define Architecture

Here is the detailed architecture that we are going to use:

1. **First convolutional layer**: consists of 32 filters with kernel_size matrix 3 by 3. Using 2-pixel strides at a time, reduce the image size by half.
2. **First pooling layer**: Using max-pooling matrix 2 by 2 (pool_size) and 2-pixel strides at a time further reduce the image size by half.
3. **Second convolutional layer**: Just like the first convolutional layer but with 16 filters only.
4. **Second pooling layer**: Same as the first pooling layer.
5. **Flattening**: Convert two-dimensional pixel values into one dimension, so that it is ready to be fed into the fully-connected layer.
6. **First dense layer + Dropout**: consists of 128 units and 1 bias unit. Dropout of rate 20% is used to prevent overfitting.
7. **Second dense layer + Dropout**: consists of 64 units and 1 bias unit. Dropout of rate 20% is also used to prevent overfitting.
8. **Output layer**: consists of only one unit and activation is a sigmoid function to convert the scores into a probability of an image being bad.

For every layer except output layer, we use Rectified Linear Unit (ReLU) activation function.

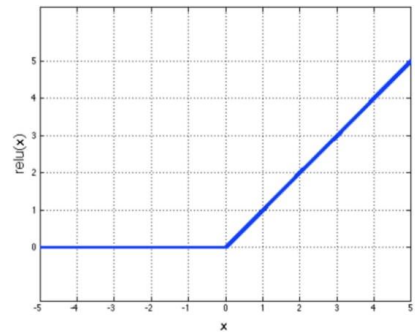

```
In [ ]:  model_cnn = Sequential(
             [
                 # First convolutional layer
                 Conv2D(filters = 32,
                        kernel_size = 3,
                        strides = 2,
                        activation = "relu",
                        input_shape = IMAGE_SIZE + (1, )),

                 # First pooling layer
                 MaxPooling2D(pool_size = 2,
                              strides = 2),

                 # Second convolutional layer
                 Conv2D(filters = 16,
                        kernel_size = 3,
                        strides = 2,
                        activation = "relu"),

                 # Second pooling layer
                 MaxPooling2D(pool_size = 2,
                              strides = 2),

                 # Flattening
                 Flatten(),

                 # Fully-connected layer
                 Dense(128, activation = "relu"),
                 Dropout(rate = 0.2),

                 Dense(64, activation = "relu"),
                 Dropout(rate = 0.2),

                 Dense(1, activation = "sigmoid")
             ]
         )

         model_cnn.summary()
```

```
Model: "sequential_1"
_________________________________________________________________
 Layer (type)                Output Shape              Param #
=================================================================
 conv2d_2 (Conv2D)           (None, 149, 149, 32)      320

 max_pooling2d_2 (MaxPooling  (None, 74, 74, 32)       0
 2D)

 conv2d_3 (Conv2D)           (None, 36, 36, 16)        4624

 max_pooling2d_3 (MaxPooling  (None, 18, 18, 16)       0
 2D)

 flatten_1 (Flatten)         (None, 5184)              0

 dense_3 (Dense)             (None, 128)               663680

 dropout_2 (Dropout)         (None, 128)               0

 dense_4 (Dense)             (None, 64)                8256

 dropout_3 (Dropout)         (None, 64)                0

 dense_5 (Dense)             (None, 1)                 65

=================================================================
Total params: 676,945
Trainable params: 676,945
Non-trainable params: 0
_________________________________________________________________
```

In [ ]: `keras.utils.plot_model(model_cnn, "card_model.png", show_shapes=True)`

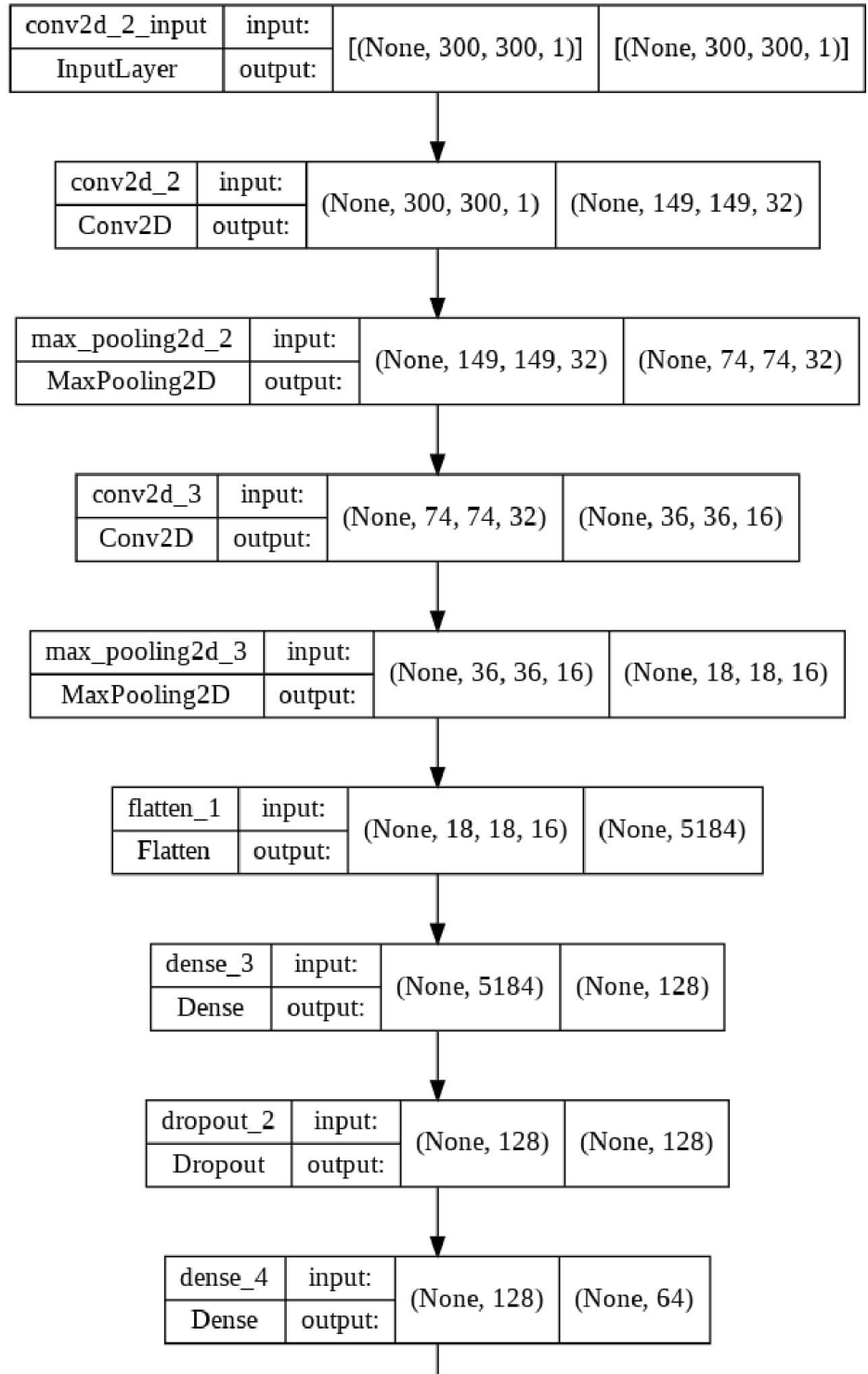

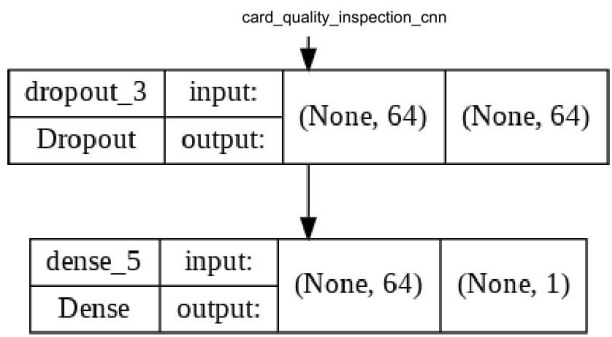

## Compile the Model

Next, we specify how the model backpropagates or update the weights after each batch feed-forward. We use `adam` optimizer and a loss function `binary cross-entropy` since we are dealing with binary classification problem. The metrics used to monitor the training progress is accuracy.

```
In [ ]: model_cnn.compile(optimizer = 'adam',
                  loss = 'binary_crossentropy',
                  metrics = ['accuracy'])
```

## Model Fitting

Before we do model fitting, let's check whether GPU is available or not.

```
In [ ]: checkpoint = ModelCheckpoint('model/card_quality_inspection_cnn_model.hdf5',
                                     verbose = 1,
                                     save_best_only = True,
                                     monitor='val_loss',
                                     mode='min')

        model_cnn.fit(train_dataset,
                      validation_data = validation_dataset,
                      batch_size = 16,
                      epochs = 20,
                      callbacks = [checkpoint],
                      verbose = 1)
```

```
Epoch 1/20
10/10 [==============================] - ETA: 0s - loss: 0.6993 - accuracy: 0.5151
Epoch 1: val_loss improved from inf to 0.68911, saving model to model/card_quality_in
spection_cnn_model.hdf5
10/10 [==============================] - 9s 831ms/step - loss: 0.6993 - accuracy: 0.5
151 - val_loss: 0.6891 - val_accuracy: 0.5068
Epoch 2/20
10/10 [==============================] - ETA: 0s - loss: 0.6941 - accuracy: 0.4832
Epoch 2: val_loss did not improve from 0.68911
10/10 [==============================] - 7s 750ms/step - loss: 0.6941 - accuracy: 0.4
832 - val_loss: 0.6936 - val_accuracy: 0.4932
Epoch 3/20
10/10 [==============================] - ETA: 0s - loss: 0.6918 - accuracy: 0.5336
Epoch 3: val_loss improved from 0.68911 to 0.68580, saving model to model/card_qualit
y_inspection_cnn_model.hdf5
10/10 [==============================] - 7s 754ms/step - loss: 0.6918 - accuracy: 0.5
336 - val_loss: 0.6858 - val_accuracy: 0.5068
Epoch 4/20
10/10 [==============================] - ETA: 0s - loss: 0.6839 - accuracy: 0.5134
Epoch 4: val_loss improved from 0.68580 to 0.67861, saving model to model/card_qualit
y_inspection_cnn_model.hdf5
10/10 [==============================] - 8s 762ms/step - loss: 0.6839 - accuracy: 0.5
134 - val_loss: 0.6786 - val_accuracy: 0.5068
Epoch 5/20
10/10 [==============================] - ETA: 0s - loss: 0.6839 - accuracy: 0.5201
Epoch 5: val_loss did not improve from 0.67861
10/10 [==============================] - 7s 733ms/step - loss: 0.6839 - accuracy: 0.5
201 - val_loss: 0.6804 - val_accuracy: 0.6486
Epoch 6/20
10/10 [==============================] - ETA: 0s - loss: 0.6874 - accuracy: 0.5906
Epoch 6: val_loss improved from 0.67861 to 0.67773, saving model to model/card_qualit
y_inspection_cnn_model.hdf5
10/10 [==============================] - 7s 750ms/step - loss: 0.6874 - accuracy: 0.5
906 - val_loss: 0.6777 - val_accuracy: 0.5068
Epoch 7/20
10/10 [==============================] - ETA: 0s - loss: 0.6842 - accuracy: 0.5369
Epoch 7: val_loss improved from 0.67773 to 0.66503, saving model to model/card_qualit
y_inspection_cnn_model.hdf5
10/10 [==============================] - 7s 798ms/step - loss: 0.6842 - accuracy: 0.5
369 - val_loss: 0.6650 - val_accuracy: 0.5203
Epoch 8/20
10/10 [==============================] - ETA: 0s - loss: 0.6797 - accuracy: 0.5906
Epoch 8: val_loss improved from 0.66503 to 0.65964, saving model to model/card_qualit
y_inspection_cnn_model.hdf5
10/10 [==============================] - 7s 746ms/step - loss: 0.6797 - accuracy: 0.5
906 - val_loss: 0.6596 - val_accuracy: 0.5068
Epoch 9/20
10/10 [==============================] - ETA: 0s - loss: 0.6737 - accuracy: 0.5839
Epoch 9: val_loss improved from 0.65964 to 0.63228, saving model to model/card_qualit
y_inspection_cnn_model.hdf5
10/10 [==============================] - 7s 748ms/step - loss: 0.6737 - accuracy: 0.5
839 - val_loss: 0.6323 - val_accuracy: 0.5338
Epoch 10/20
10/10 [==============================] - ETA: 0s - loss: 0.6510 - accuracy: 0.6191
Epoch 10: val_loss improved from 0.63228 to 0.59559, saving model to model/card_quali
ty_inspection_cnn_model.hdf5
10/10 [==============================] - 9s 848ms/step - loss: 0.6510 - accuracy: 0.6
191 - val_loss: 0.5956 - val_accuracy: 0.7905
Epoch 11/20
10/10 [==============================] - ETA: 0s - loss: 0.6296 - accuracy: 0.6879
```

```
Epoch 11: val_loss improved from 0.59559 to 0.53317, saving model to model/card_quali
ty_inspection_cnn_model.hdf5
10/10 [==============================] - 7s 756ms/step - loss: 0.6296 - accuracy: 0.6
879 - val_loss: 0.5332 - val_accuracy: 0.7770
Epoch 12/20
10/10 [==============================] - ETA: 0s - loss: 0.6102 - accuracy: 0.6795
Epoch 12: val_loss improved from 0.53317 to 0.48563, saving model to model/card_quali
ty_inspection_cnn_model.hdf5
10/10 [==============================] - 7s 756ms/step - loss: 0.6102 - accuracy: 0.6
795 - val_loss: 0.4856 - val_accuracy: 0.8514
Epoch 13/20
10/10 [==============================] - ETA: 0s - loss: 0.5905 - accuracy: 0.7013
Epoch 13: val_loss improved from 0.48563 to 0.48004, saving model to model/card_quali
ty_inspection_cnn_model.hdf5
10/10 [==============================] - 7s 754ms/step - loss: 0.5905 - accuracy: 0.7
013 - val_loss: 0.4800 - val_accuracy: 0.8649
Epoch 14/20
10/10 [==============================] - ETA: 0s - loss: 0.5886 - accuracy: 0.7081
Epoch 14: val_loss improved from 0.48004 to 0.41873, saving model to model/card_quali
ty_inspection_cnn_model.hdf5
10/10 [==============================] - 7s 756ms/step - loss: 0.5886 - accuracy: 0.7
081 - val_loss: 0.4187 - val_accuracy: 0.8784
Epoch 15/20
10/10 [==============================] - ETA: 0s - loss: 0.5284 - accuracy: 0.7433
Epoch 15: val_loss did not improve from 0.41873
10/10 [==============================] - 7s 742ms/step - loss: 0.5284 - accuracy: 0.7
433 - val_loss: 0.4318 - val_accuracy: 0.8378
Epoch 16/20
10/10 [==============================] - ETA: 0s - loss: 0.5545 - accuracy: 0.7131
Epoch 16: val_loss improved from 0.41873 to 0.40901, saving model to model/card_quali
ty_inspection_cnn_model.hdf5
10/10 [==============================] - 7s 752ms/step - loss: 0.5545 - accuracy: 0.7
131 - val_loss: 0.4090 - val_accuracy: 0.8108
Epoch 17/20
10/10 [==============================] - ETA: 0s - loss: 0.5137 - accuracy: 0.7567
Epoch 17: val_loss improved from 0.40901 to 0.40142, saving model to model/card_quali
ty_inspection_cnn_model.hdf5
10/10 [==============================] - 8s 763ms/step - loss: 0.5137 - accuracy: 0.7
567 - val_loss: 0.4014 - val_accuracy: 0.7905
Epoch 18/20
10/10 [==============================] - ETA: 0s - loss: 0.4644 - accuracy: 0.7836
Epoch 18: val_loss improved from 0.40142 to 0.34116, saving model to model/card_quali
ty_inspection_cnn_model.hdf5
10/10 [==============================] - 8s 768ms/step - loss: 0.4644 - accuracy: 0.7
836 - val_loss: 0.3412 - val_accuracy: 0.8446
Epoch 19/20
10/10 [==============================] - ETA: 0s - loss: 0.4579 - accuracy: 0.8087
Epoch 19: val_loss improved from 0.34116 to 0.33842, saving model to model/card_quali
ty_inspection_cnn_model.hdf5
10/10 [==============================] - 7s 757ms/step - loss: 0.4579 - accuracy: 0.8
087 - val_loss: 0.3384 - val_accuracy: 0.8446
Epoch 20/20
10/10 [==============================] - ETA: 0s - loss: 0.4704 - accuracy: 0.7685
Epoch 20: val_loss improved from 0.33842 to 0.31583, saving model to model/card_quali
ty_inspection_cnn_model.hdf5
10/10 [==============================] - 7s 750ms/step - loss: 0.4704 - accuracy: 0.7
685 - val_loss: 0.3158 - val_accuracy: 0.8581
```

Out[ ]:  `<keras.callbacks.History at 0x7f2ebca1bb90>`

## Training Evaluation

Let's plot both loss and accuracy metrics for train and validation data based on each epoch.

```python
plt.subplots(figsize = (8, 6))
sns.lineplot(data = pd.DataFrame(model_cnn.history.history,
                                 index = range(1, 1+len(model_cnn.history.epoch))))
plt.title("TRAINING EVALUATION", fontweight = "bold", fontsize = 20)
plt.xlabel("Epochs")
plt.ylabel("Metrics")

plt.legend(labels = ['val loss', 'val accuracy', 'train loss', 'train accuracy'])
plt.show()
```

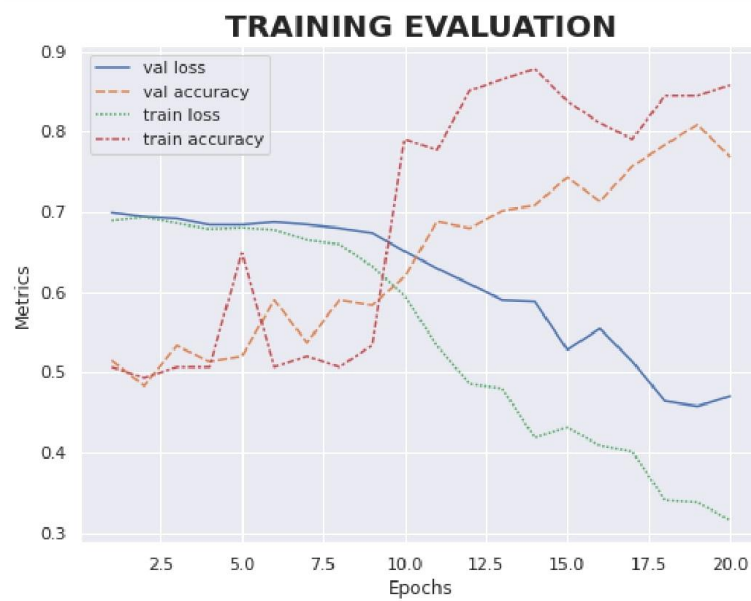

We can conclude that the model is **overfitting** the data since both train loss and val loss simultaneously dropped towards zero. Also, both train accuracy and val accuracy increase towards 100%.

## Testing on Unseen Images

Our model performs very well on the training and validation dataset which uses augmented images. Now, we test our model performance with unseen and unaugmented images.

```python
best_model = load_model("model/card_quality_inspection_cnn_model.hdf5")
```

```python
y_pred_prob = best_model.predict(test_dataset)
```

The output of the prediction is in the form of probability. We use THRESHOLD = 0.5 to separate the classes. If the probability is greater or equal to the THRESHOLD, then it will be classified as bad, otherwise ok.

```python
THRESHOLD = 0.5
y_pred_class = (y_pred_prob >= THRESHOLD).reshape(-1,)
y_true_class = test_dataset.classes[test_dataset.index_array]

pd.DataFrame(
    confusion_matrix(y_true_class, y_pred_class),
    index = [["Actual", "Actual"], ["ok", "bad"]],
    columns = [["Predicted", "Predicted"], ["ok", "bad"]],
)
```

Out[ ]:
|  |  | Predicted | |
|---|---|---|---|
|  |  | ok | bad |
| Actual | ok | 94 | 1 |
|  | bad | 0 | 91 |

```python
print(classification_report(y_true_class, y_pred_class, digits = 4))
```

```
              precision    recall  f1-score   support

           0     1.0000    0.9895    0.9947        95
           1     0.9891    1.0000    0.9945        91

    accuracy                         0.9946       186
   macro avg     0.9946    0.9947    0.9946       186
weighted avg    0.9947    0.9946    0.9946       186
```

According to the problem statement, we want to minimize the case of False Negative, where the bad part is misclassified as `ok`. This can cause the whole order to be rejected and create a big loss for the company. Therefore, in this case, we prioritize Recall over Precision.

But if we take into account the cost of re-plating a product, we have to minimize the case of False Positive also, where the ok product is misclassified as `bad`. Therefore we can prioritize the `F1 score` which combines both Recall and Precision.

On test dataset, the model achieves a very good result as follow:

- Accuracy: 99.46%
- Recall: 98.95%
- Precision: 98.91%
- F1 score: 99.46%

---

# Conclusion

By using CNN and on-the-fly data augmentation, the performance of our model in training, validation, and test images is almost perfect, reaching 99.46% accuracy and F1 score. We can utilize this model by embedding it into a surveillance camera where the system can automatically separate bad product from the production line. This method surely can reduce human error and human resources on manual inspection, but it still needs supervision from human since the model is not 100% correct at all times.

```
In [ ]:
```